# Viewpoint
# A Theoretical Computer Science Perspective on Consciousness and Artificial General Intelligence[1]

Lenore Blum[2] and Manuel Blum[3]

## Abstract

We have defined the **Conscious Turing Machine (CTM)** for the purpose of investigating a **Theoretical Computer Science (TCS)** approach to consciousness. For this, we have hewn to the TCS demand for simplicity and understandability. The CTM is consequently and intentionally a simple machine. It is not a model of the brain, though its design has greatly benefited - and continues to benefit - from neuroscience and psychology. The CTM is a model of and for consciousness.

Although it is developed to understand consciousness, the CTM offers a thoughtful and novel guide to the creation of an **Artificial General Intelligence (AGI)**. For example, the CTM has an enormous number of powerful processors, some with specialized expertise, others unspecialized but poised to develop an expertise. For whatever problem must be dealt with, the CTM has an excellent way to utilize those processors that have the required knowledge, ability, and time to work on the problem, even if it is not aware of which ones these may be.

## 1    The Conscious Turing Machine (CTM) in a nutshell

A **Theoretical Computer Science (TCS)** perspective is employed to define the **Conscious Turing Machine (CTM)**, "conscious awareness", and the "feeling of consciousness" in the CTM (Blum & Blum, 2021). These are followed by arguments explaining *why* the definitions capture commonly accepted *understandings* of consciousness and the *feelings* that many people have of their own consciousness (Blum & Blum, 2022).

The CTM is a mathematical formalization of a modified version of the **Global Workspace Theory (GWT)** of consciousness (Figure 1) that originated with cognitive neuroscientist Bernard Baars in (Baars B. J., 1988) and (Baars B. J., 1997), and was subsequently extended to the Global Neuronal Workspace (GNW) by (Dehaene & Changeux, 2011), (Dehaene S. , 2014) and (Mashour, Roelfsema, Changeux, & Dehaene, 2020).[4]

Baars describes conscious awareness through a theater analogy as the activity of actors in a play performing on a stage of Working Memory, their performance under observation by a huge audience of unconscious processors sitting in the dark.

In the CTM, the stage is represented by a **Short Term Memory (STM)** that at every tick of a central clock contains what is defined to be CTM's conscious content. The audience members are represented by an enormous collection of powerful random access processors - some with their own expertise, some without, and all with deep learning hardware to develop or improve an expertise - that make up CTM's **Long Term Memory (LTM)** processors. These LTM processors make predictions about the world and get feedback from that (CTM's) world. Based on that feedback, learning algorithms Internal to each processor improve that processor's behavior.

---

[1] This work was supported in part by a grant from UniDT.

[2] lblum@cs.cmu.edu; lenore.blum@berkeley.edu

[3] mblum@cs.cmu.edu; blum@cs.berkeley.edu

[4] Baars's GWT is strongly influenced by earlier work in cognitive science, much of which was done at Carnegie Mellon: (Simon, 1969), (Reddy, 1976), (Newell, 1990) and (Anderson, 1996).

LTM processors compete to get their questions, answers, and comments onto the STM stage for immediate broadcast to the audience. The information that passes through STM is coded in the form of chunks. *Conscious awareness/attention* is defined formally in the CTM as the reception by all LTM processors of whatever chunk was broadcast from STM and received by processors of LTM. In time, some LTM processors become connected via links that serve as channels for carrying chunks directly between processors. Links turn *indirect* conscious communication (via STM) between LTM processors into *direct* unconscious communication (not involving STM) between those processors.

While these definitions are natural, they are merely definitions. They are not arguments that the CTM is conscious in the sense that the word "consciousness" is normally used. We do argue however that the definitions and explanations from the CTM capture broadly accepted intuitive understandings of consciousness.

Although inspired by Baars' global workspace model, there are significant differences between Baars' model (Figure 1a) and the CTM (Figure 1b). With respect to architecture, while Baars has a central executive, the CTM has none: it is a distributed system that enables the emergence of functionality and applications to general intelligence. In the CTM, input sensors transmit environmental information *directly to* appropriate LTM processors; output actuators act on the environment based on information gotten *directly* from specific LTM processors. In the Baars model, these inputs and outputs are processed through working memory. In the CTM, chunks are *formally defined* and are submitted by LTM processors to compete in a *well-defined competition* for STM; in the Baars model, neither are formally defined. For Baars, a conscious event occurs between the input and the central executive; in the CTM, conscious awareness is the reception by the LTM processors of the chunk broadcast globally from the STM.

Although inspired by Turing's simple yet powerful model of a computer, the CTM is *not* a standard Turing Machine. That's because what gives the CTM a "feeling of consciousness" is not its input-output map or its computing power but rather its *global workspace architecture;* its *predictive dynamics* (cycles of prediction, feedback, and learning); *its rich multi-modal inner language* (which we call Brainish) for inter-processor communication; and certain particularly important LTM *processors* including the Inner Speech, Inner Generalized Sensation, and *Model-of-the-World* processors.

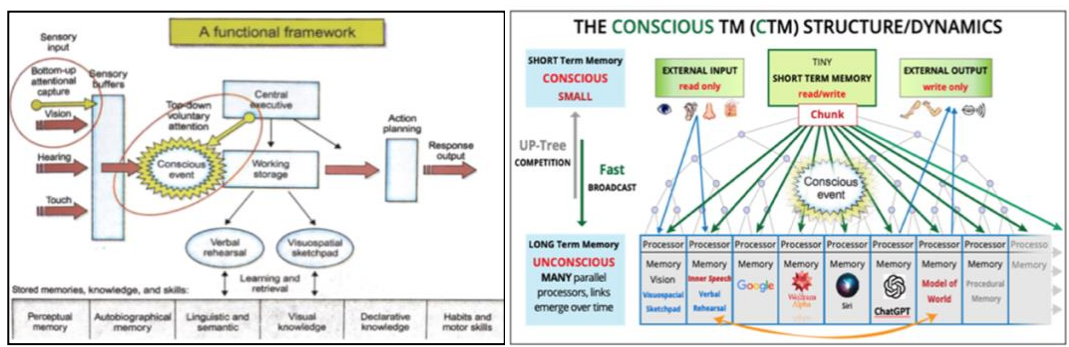

Figure 1. Sketches of models: a) Baars' GWT model (left) and b) the CTM (right).

As mentioned in the abstract, the CTM is not intended to be a model of the brain but a simple model of consciousness, and even there, the CTM model can hardly be expected to explain everything: it is too simple for that. The reasonableness of the model (and its TCS perspective) should be judged by its contribution to the discussion and understanding of consciousness, to related topics like feelings of pain and pleasure, and to its potential for use as an AGI.

The above presents an *overview* of the CTM model. We *refer* the reader to two papers for the formal definition of CTM as 7-tuple and chunk as 6-tuple. The first paper (Blum & Blum, 2021) explores explanations for the feelings of pain and pleasure in the CTM.[5] The second paper (Blum & Blum, 2022) explores additional phenomena generally associated with consciousness such as free will and the disorder of blindsight.[6] We give explanations *derived from the formal model* and draw confirmation from consistencies *at a high-level* with the psychology and neuroscience literature.[7,8]

## 2    The CTM and Artificial General Intelligence (AGI)[9]

Though the CTM is defined to be a very simple model of consciousness, it being explicitly formally defined for generating definitions and understandings of consciousness, it also suggests a novel approach to AGI, giving a way to coordinate an enormous number of (special-purpose) artificial intelligence (AI) agents for the purpose of building the AGI. In particular, it suggests how to coordinate a huge number - $10^7$ or more[10] - of processors, some specialized, most initially unspecialized but capable of being specialized, to solve a variety of unforeseen problems. In an AGI, specialized processors could be tasked to get information from a number of search engines, from ChatGPT or GPT-4, Wikipedia, Google Translate, Wolfram alpha, the weather channel, newspapers, HOL Light[11], and so on. These are existing ready-made processors. Many more processors could and would be developed from scratch as needed by the CTM itself.

**A principal contribution of the CTM is a way to coordinate processors that must solve a diverse collection of unforeseen problems.** The CTM assigns tasks to its processors, even though it has no

---

[5] For an update on pain and pleasure in the CTM, see Chapter 4 of the Blum's *The Hard Problem for Pain and Pleasure*, in https://arxiv.org/abs/2011.09850.

[6] For an update see https://arxiv.org/abs/2107.13704.

[7] We note a historical synergy between theoretical computer science and neuroscience. Turing's simple computer model led neuroscientist Warren S. McCulloch and mathematician Walter Pitts to define their formal neuron, itself a simple model of a neuron (McCulloch & Pitts, 1943). Mathematics forced their model to have inhibition, not just excitation - because without inhibition, loop-free circuits of formal neurons can only compute monotonic functions - and these do not suffice to build a universal Turing Machine. The McCulloch-Pitts neuron also gave rise to the mathematical formalization of neural nets (Wikipedia) and subsequent deep learning algorithms (Goodfellow, Bengio, & Courville, 2016), further illustrating ongoing synergies.

[8] A forthcoming monograph (Blum, Blum, & Blum, In preparation) describes in more detail how the CTM works. Its three appendices demonstrate how CTM can operate with *no* central executive, despite that all other global workspace models (such as Baars' functional model (Baars B. J., 1997)) hypothesize a Central Executive. These same appendices show by example how CTM functions with just one chunk in STM instead of George Miller's 7±2 (Miller, 1956) or Nelson Cowan's 3 or 4 (Cowan, 2015). No other models suggest that one chunk will suffice.

[9] **Artificial General Intelligence** (**AGI**) is the ability of an intelligent agent to understand or learn any intellectual task that can be learned by human beings or other animals.

[10] $10^7$ is the estimated number of cortical columns in the brain.

[11] HOL Light is a formal mathematical programming system for generating proofs that are logically and mathematically correct, and/or for checking any "proofs" given it. It was used, for example, by Tom Hales to prove the correctness of his solution to the Kepler conjecture (Hales, 2005).

central executive and no single processor or collection of processors to keep track of which processors have the time and know-how to do the task. How it does that is an interesting koan.

Suppose (a processor of) CTM has a task to perform but nary a clue how to perform it, and no idea which, if any, of its many **Long Term Memory (LTM) processors** has the knowledge, ability, and time to deal with the task. Through a well-defined competition for **Short Term Memory (STM)**, the processor submits a request for help to *all* LTM processors. The request will with some probability **[**well-defined in the CTM**]** reach STM for global broadcast to all (LTM) processors. All processors that have relevant expertise and time to work on the problem respond, again through the competition and global broadcast. Their broadcasts in turn can motivate other processors to come into play. In this way, the CTM engages powerful processors to collectively solve a problem that CTM had no idea how to solve, no approach to the problem, no sense which processors, if any, could be helpful.

**When it comes to mathematics**, the CTM is ideal for orchestrating its processors to recognize a sound logical argument, write a correct mathematical proof, and check its work. It can program its processors and modify them as needed to reach such goals. For example, one processor can suggest an approach to a proof, a second can evaluate the likelihood that the approach will work, a third can outline a potential "proof", a fourth can check if a proposed "proof" is really a proof (pointing out what problems arise) if not, and so on...

**More generally, the CTM can and must have processors for checking the truth** of statements or arguments. Consider for example the assertion that shrimp is healthy to eat. One source says YES, shrimp is healthy, but that statement comes from an Association that represents the Frozen Foods industry, making it suspect. Another says NO, shrimp is unhealthy: it has lots of cholesterol and cholesterol is unhealthy. Yet another says YES, shrimp is healthy, the cholesterol in shrimp is the healthy LDL kind. As that last paper (De Oliveira e Silva, et al., 1996) is authored by a scholar from a respected (Rockefeller) University, and published in a reputable refereed journal, its case is the strongest so far. Responses to the paper may further strengthen or weaken the assessment.

## 3   What features does the CTM bring to the design of an AGI?

The CTM is a simple TCS model of consciousness. It is not a brain. It is not an AGI. That said, we suggest that its basic features surely have value in the design of some aspects of an AGI. For example:

   **3.1.** The CTM suggests an approach to building an AGI that has **no central executive** - no conductor, no stage director. It has an enormous number of processors, each of which is largely self-directing, rather like the members of a self-conducting musical ensemble. This architecture is unexpected and strange because large assemblies, large orchestras, and large political states, generally have a leader.

The CTM has one and only one actor on stage, and that one is *not* a leader. It serves merely as
        **3.1.1.** a small buffer to hold the winning chunk of the current competition and
        **3.1.2.** a broadcasting station to beam that chunk to the entire LTM audience.

The CTM solves a conundrum: how is it possible for a long and subtle argument, say the proof of a difficult theorem, to be understood – grasped as it were in the palm of one's hand? That handful is the final chunk that contains something like "Eureka! I got it." That chunk is from a processor that, if asked, can point to the outline of a proof, each phrase of which can point to what in the proof it depends on and what depends on it.

**3.2.** The audience members of the CTM global workspace are **self-monitoring processors**. They have the final word on the value of their personal contribution.[12]

**3.3.** Baar**s** (Baars B. J., 1997) says that the audience members of the global workspace consult among themselves to agree on who to send to the stage, but how do they do that? Baars doesn't say.

The CTM, on the other hand, explains precisely how to do it. It hosts a **well-defined competition** that is similar to, and actually provably better than a chess or tennis tournament in that, at minor cost and negligible extra time, it *guarantees* that each processor will broadcast its information with probability proportional to the value of its information (a value computed dispassionately by the processor's sleeping experts algorithm), something that chess and tennis tournaments do not and cannot do.

**3.4. Sleeping Experts** learning algorithms (Blum A. , 1997), (Blum, Hopcroft, & Kannan, 2015) determine how a processor assigns a value to its information, a value that is (mostly) self-determined by the processor. The CTM makes do with no teacher at the head of the class and no answer sheet to make corrections. Its processors self-predict and self-correct based on feedback. We expect AGI designers to be interested in how they manage that.

**3.5. CTM's Model-of-the-World (MotW)** processor develops **world models**. These world models are hugely important for planning, testing, making corrections, distinguishing fiction from nonfiction, living from non-living, self from not-self, and most importantly for contributing to feelings of consciousness. The CTM has at birth a rudimentary MotW Processor, then continuously upgrades it and its models. How the CTM creates and manages its world models is especially important given that the CTM does not consciously see the world directly, as does the Baars model (Figure 1a), but indirectly through its world models (Figure 1b).

**3.6. The CTM can explain what it is doing and why**. It can answer questions about the how and why of its doings and give arguments to support its answers.

# 4 Arguments for and against using CTM as a guide for creating an AGI

The specification of CTM[13] gives a sense of how a CTM works. Descriptions are given of how each processor assigns a valenced measure of importance (a weight) to its chunks, and how that measure is affected by the Sleeping Experts Algorithm in each LTM processor. There is a description of how the tournament that starts at time t is run, that tournament being a competition among all N chunks[14] contributed by all N processors at time t. Each such tournament takes ($\log_2 N$) steps: the first step being N/2 matches performed in parallel, the second being N/4, ..., the last being a single match to crown the winner. The tournament is as fast as any tournament for tennis and chess but better as chess and tennis tournaments don't guarantee, as does the CTM tournament, that chunks get to STM with probability proportional to their importance. On that account, CTM processors can remain hard-wired and in place without in any way affecting which chunk will win any given tournament.

Another argument for using CTM as a guide for AGI comes from neuroscience research demonstrating that in humans, "language and thought are not the same thing" (Fedorenko & Varley, 2016). Individuals with global aphasia, "despite their near-total loss of language are nonetheless able to add and subtract, solve logic problems, think about another's thoughts, appreciate music, ...." Healthy adults "strongly engage the brain's language areas when they understand a sentence, but not when they perform non-linguistic tasks such as arithmetic, storing information in working memory, ..., or listening to music."

---

[12] The idea reminds us of the visiting Admiral at MIT who told McCulloch's neurophysiology group in 1959 that in the navy, it is not the flagship that commands the fleet: it is the ship with the information. The CTM's processors are the ships of an enormous fleet. The workings of the CTM give precise meaning to the Admiral's words.

[13] In (Blum & Blum, 2021), (Blum & Blum, 2022), and in the upcoming monograph (Blum, Blum, & Blum, In preparation).

[14] We assume that N = $2^k$ for some positive integer k.

Influenced by this research and comparing large language models to formal and functional properties of human language, (Mahowald, et al., 2023) argue that while large language models "are good models of language", [they are] "incomplete models of human thought." They further argue "that future language models can master both formal and functional linguistic competence by establishing a division of labor between the core language system and components for other cognitive processes, ..." as in the human brain. They provide two suggestions to accomplish this. Their first suggestion is **Architectural Modularity** (whereby separate specialized modules work together with each other). The CTM incorporates such modularity by utilizing multiple processors with different input domains, knowledge, and functionalities.

Their second suggestion, **Emergent Modularity** (modularity that emerges within a large language model), points to the possibility that deep learning alone will suffice for AGI, though they argue that architectural modularity is "much better aligned with... real-life language use".

The possibility of emergence is supported by (Bubeck, et al., 2023) who examine the impressive and multiple "sparks" of general intelligence demonstrated by early experiments with the large language model GPT-4 and view it as an early version of an AGI.

Indeed, it may turn out that no global workspace model or CTM is needed to create an AGI, that deep learning alone suffices, that a single machine with a sufficiently large matrix size can be a universal AGI, but we doubt it. One can argue that the matrix size of a deep learning AGI must grow with the square of the number of problems it is to solve, and such a size would be difficult to achieve since the best current AIs currently use about $10^{14}$ parameters. The CTM, which is designed for understanding consciousness, can reasonably handle $10^7$ AIs with $10^{14}$ parameters per AI, for a total of $10^{21}$ parameters. For comparison, there are $10^{11}$ stars in the milky way galaxy and $2 \times 10^{23}$ stars in the visible universe. Avogadro's number is 3 time as large as that at ≈ $6.0221 \times 10^{23}$.

Returning to consciousness, the CTM global workspace model is a promising untapped approach to turning AI into AGI. We expect that robots with CTM-like brains that construct models of the world will have "feelings of consciousness", hence be more likely to experience empathy. Finally, as AIs become more human-like, understanding consciousness and feelings of pain will be critical if we want to avoid inflicting suffering on our planet's co-inhabitants.

### Acknowledgements

The work of Lenore Blum and Manuel Blum was supported in part by Carnegie Mellon University (CMU), in part by a sabbatical year from CMU at the Simon's Institute for the Theory of Computing, and in part by a generous gift from UniDT. We are grateful to Jean-Louis Villecroze for his ongoing work to simulate CTM, Paul Liang for his insight into multimodal Brainish (Liang, 2022), for our students at CMU and Peking U who constantly challenge us, and our friends and colleagues Raj Reddy and Michael Xuan for their suggestions, personal support, and extraordinary encouragement.

## References

Anderson, J. R. (1996). ACT: A simple theory of complex cognition. *American Psychologist, 51(4)*, 355-365.

Baars, B. J. (1988). *A Cognitive Theory of Consciousness.* Cambridge: Cambridge University Press.

Baars, B. J. (1997). In the Theater of Consciousness: A rigorous scientific theory of consciousness. *Journal of Consciousness Studies 4, No. 4*, 292-309.

Blum, A. (1997). Empirical support for winnow and weighted-majority algorithms: Results on a calendar scheduling domain. *Machine Learning, 26*(1), 5-23.

Blum, A., Hopcroft, J., & Kannan, R. (2015). *Foundations of Data Science.* Ithaca. Retrieved from https://www.cs.cornell.edu/jeh/book.pdf

Blum, L., & Blum, M. (2022, May 24). A theory of consciousness from a theoretical computer science perspective: Insights from the Conscious Turing Machine. *PNAS, 119*(21).

Blum, M., & Blum, L. (2021, March). A Theoretical Computer Science Perspective on Consciousness. *JAIC, 8*(1), 1-42. https://www.worldscientific.com/doi/epdf/10.1142/S2705078521500028.

Blum, M., Blum, L., & Blum, A. (In preparation). *Towards a Conscious AI: A Computer Architecture Inspired by Cognitive Neuroscience.*

Bubeck, S., Chandrasekaran, V., Eldan, R., Gehrke, J., Horvitz, E., Kamar, E., . . . Zhang, Y. (2023, March 22). *Sparks of Artificial General Intelligence: Early experiments with GPT-4.* Retrieved from arXiv: https://arxiv.org/abs/2303.12712

Cowan, N. (2015, July). George Miller's magical number of immediate memory in retrospect: Observations on the faltering progression of science. *Psychol Rev., 122*(3), 536-41.

De Oliveira e Silva, E. R., Seidman, C. E., Tian, J. J., Hudgins, L. C., Sacks, F. M., & Breslow, J. L. (1996, November ). Effects of shrimp consumption on plasma lipoproteins. *The American Journal of Clinical Nutrition, 64*(6), 712- 717, https://doi.org/10.1093.

Dehaene, S. (2014). *Consciousness and the Brain: Deciphering How the Brain Codes Our Thoughts.* New York: Viking Press.

Dehaene, S., & Changeux, J. P. (2011). Experimental and theoretical approaches to conscious processing. *Neuron; 70(2)*, 200-227.

Fedorenko, E., & Varley, R. (2016, April). Language and thought are not the same thing: Evidence from neuroimaging and neurological patients. *Annals of the New York Academy of Sciences, 1369* , 132-153, doi:10.1111/nyas.13046.

Goodfellow, I., Bengio, Y., & Courville, A. (2016). *Deep Learning.* Cambridge, MA: MIT Press.

Hales, T. (2005). A proof of the Kepler conjecture. *Annals of Mathematics, 162*, 1065–1185.

Liang, P. P. (2022). *Brainish: Formalizing A Multimodal Language for Intelligence and Consciousness.* Retrieved from arXiv: https://doi.org/10.48550/arXiv.2205.00001

Mahowald, K., Ivanova, A. A., Blank, I. A., Kanwisher, N., Tenenbaum, J. B., & Fedorenko, E. (2023, January 18). *Dissociating language and thought in large language models: a cognitive perspective.* Retrieved from arXiv: https://arxiv.org/abs/2301.06627

Mashour, G. A., Roelfsema, P., Changeux, J.-P., & Dehaene, S. C. (2020). Conscious Processing and the Global Neuronal Workspace Hypothesis. *Neuron, 195*(5), 776-798.

McCulloch, W. S., & Pitts, W. (1943). A logical calculus of the ideas immanent in nervous activity. *A logical calculus of the ideas immanent in nervous activity, 5*, 115-133.

Miller, G. A. (1956). The magical number seven, plus or minus two: Some limits on our capacity for processing information. *Psychological Review, 63*, 81-97.

Newell, A. (1990). *Unified Theories of Cognition.* Cambridge: Harvard University Press.

Reddy, D. R. (1976, April). Speech Rcogniton by Machine: A Review. *Proceedings of the IEEE*, 501-531. Retrieved from http://www.rr.cs.cmu.edu/sr.pdf

Simon, H. A. (1969). *The Sciences of the Artificial.* Cambridge, MA, USA: MIT Press.

Wikipedia. (n.d.). *History of artificial neural networks*. Retrieved from https://en.wikipedia.org/wiki/History_of_artificial_neural_networks